\pdfoutput=1
\documentclass[11pt]{article}

\usepackage{acl}

\usepackage{times}
\usepackage{latexsym}
\usepackage{caption}
\usepackage{subcaption}
\usepackage{multirow}
\usepackage[utf8]{inputenc}
\usepackage{a4wide}
\usepackage{placeins}
\usepackage{setspace}
\usepackage{booktabs}
\usepackage{rotating}
\usepackage{lscape}
\usepackage{color}
\usepackage{longtable}
\usepackage{rotating}
\usepackage{comment}
\usepackage{hyperref}
\usepackage{graphicx}
\usepackage{dblfloatfix}
\usepackage{amsmath}
\usepackage[T1]{fontenc}
\usepackage[utf8]{inputenc}

\usepackage{microtype}

\title{Measuring the Impact of (Psycho-)Linguistic and Readability Features and Their Spill Over Effects on the Prediction of Eye Movement Patterns}

\author{Daniel Wiechmann\\
  University of Amsterdam\\
  \texttt{\normalsize d.wiechmann@uva.nl} \\\And
  Yu Qiao \\
  RWTH-Aachen University\\
  \texttt{\normalsize yu.qiao@rwth-aachen.de} \\
  \AND
  Elma Kerz\\
  RWTH-Aachen University\\
  \texttt{\normalsize elma.kerz@ifaar.rwth-aachen.de} \\\And
  Justus Mattern\\
  RWTH-Aachen University\\
  \texttt{\normalsize justus.mattern@rwth-aachen.de}
  }

\begin{document}
\maketitle
\begin{abstract}
There is a growing interest in the combined use of NLP and machine learning methods to predict gaze patterns during naturalistic reading. While promising results have been obtained through the use of transformer-based language models, little work has been undertaken to relate the performance of such models to general text characteristics. In this paper we report on experiments with two eye-tracking corpora of naturalistic reading and two language models (BERT and GPT-2). In all experiments, we test effects of a broad spectrum of features for predicting human reading behavior that fall into five categories (syntactic complexity, lexical richness, register-based multiword combinations, readability and psycholinguistic word properties). Our experiments show that both the features included and the architecture of the transformer-based language models play a role in predicting multiple eye-tracking measures during naturalistic reading.  We also report the results of experiments aimed at determining the relative importance of features from different groups using SP-LIME. 
\end{abstract}

\section{Introduction}

Extensive studies using eye-trackers to observe gaze patterns have shown that humans read sentences efficiently by performing a series of fixations and saccades (for comprehensive overviews, see, e.g. \citet{rayner2012psychology}, \citet{seidenberg2017language}, and \citet{brysbaert2019many}). During a fixation, the eyes stay fixed on a word and remain fairly static for 200-250 milliseconds. Saccades are rapid jumps between fixations that typically last 20-40 ms and span 7-9 characters. In addition, when reading, humans do not fixate one word at a time, i.e. some saccades run in the opposite direction, and some words or word combinations are fixed more than once or skipped altogether. Much of the early work in this area was concerned with the careful construction of sentences to model human reading behavior and understand predictive language processing \cite{staub2015effect,kuperberg2016we}. The use of isolated, decontextualized sentences in human language processing research has been questioned on ecological validity grounds. With the growing awareness of the importance of capturing naturalistic reading, new corpora of eye movement data over contiguous text segments have emerged. Such corpora serve as a valuable source of data for establishing the basic benchmarks of eye movements in reading and provide an essential testing ground for models of eye movements in reading, such as the E-Z Reader model \cite{reichle1998toward} and the SWIFT model \cite{engbert2005swift}. They are also used to evaluate theories of human language processing in psycholinguistics: For example, the predictions of two theories of syntactic processing complexity (dependency locality theory and surprisal) were tested in the Dundee Corpus, which contains the eye-tracking record of 10 participants reading 51,000 words of newspaper text \cite{demberg2008data}. Subsequent work has presented accounts where the ability of a language model to predict reading times is a linear function of its perplexity \cite{goodkind2018predictive}. More recent work has employed transformer-based language models to directly predict human reading patterns across new datasets of eye-tracking and electroencephalography during natural reading \citep[][for more details see the related work section below]{schrimpf2020neural, hollenstein-etal-2021-multilingual}. While this work has made significant progress, there is limited work aimed at determining the role of general text properties in predicting eye movement patterns in corpora of naturalistic reading. To date, research has addressed this issue only peripherally \cite{lowder2018lexical,snell2020story,hollenstein-etal-2021-multilingual}, examining the role of text features only on the basis of a small number of linguistic features. 

In this paper, we conduct a systematic investigation of the effects of text properties on eye movement prediction: We determine the extent to which these properties affect the prediction accuracy of two transformer-based language models, BERT and GPT-2. The relationship between these properties and model performance is investigated in two ways: (a) building on the approaches in \citet{lowder2018lexical} and \citet{hollenstein-etal-2021-multilingual}, by investigating the sensitivity of model predictions to a wide range of text features, and (b) by incorporating text features into the transformer-based language models. With respect to the latter, we examine the effects of the preceding sentence on gaze measurement within the sentence of interest. This was motivated by psycholinguistic literature that has demonstrated ``spillover'' effects, where the fixation duration on a word is affected by linguistic features of the preceding context \citep[][see also \citet{barrett2020sequence} for a reference to the utility of information about preceding input]{pollatsek2008immediate,shvartsman2014computationally}.  Computational reading models have not addressed linguistic concepts beyond the level of the fixated word much, with a few exceptions, e.g. spillover effects related to previewing the next word n+1 during the current fixation on word n \cite{engbert2005swift}. Here we extend the study of spillover effects to the effects of textual features of the preceding sentence. To our knowledge, this is the first systematic attempt to investigate the effects of textual features on the prediction of eye-tracking measures in a corpus of naturalistic reading by considering a large number of features spanning different levels of linguistic analysis.

\section{Related work}

In this section, we provide a brief overview of the available literature that has used transformer-based language models to predict human reading patterns, as well as the literature that has investigated the role of text properties on word predictability during naturalistic reading.

\citet{schrimpf2020neural} evaluated a broad range of language models on the match of their internal representations to three datasets of human neural activity (fMRI and ECoG) during reading. Their results indicated that transformer-based models perform better than recurrent networks or word-level embedding models. They also found that the models with the best match with human language processing were models with unidirectional attention transformer architectures: specifically the generative pretrained transformer (GPT-2) \cite{radford2019language}, consistently outperformed all other models in both fMRI and ECoG data from sentence-processing tasks. \citet{hollenstein-etal-2021-multilingual} presented the first study analyzing to what extent transformer language models are able to directly predict human gaze patterns during naturalistic reading. They compare the performance of language-specific and multilingual pretrained and fine-tuned BERT and XLM models to predict reading time measures of eye-tracking datasets in four languages (English, Dutch, German, and Russian). Their results show that both monolingual and multilingual transformer-based models achieve surprisingly high accuracy in predicting a range of eye-tracking features across all four languages. For the English GECO dataset, which is also used in the current study, the BERT and XLM models yielded prediction accuracies (100 - mean absolute error (MAE)) ranging between 91.15\% (BERT-EN) and 93.89\% (XLM-ENDE). 

To our knowledge, the first study to investigate the role of textual characteristics on word predictability during naturalistic reading is an experimental study conducted by \citet{lowder2018lexical}. This study implemented a large-scale cumulative cloze task to collect word-by-word predictability data (surprisal and entropy reduction scores) for 40 text passages which were subsequently read by 32 participants while their eye movements were recorded. \citet{lowder2018lexical} found that surprisal scores were associated with increased reading times in all eye-tracking measures. They also observed a significant effect of text difficulty, measured by Flesch–Kincaid grade level of each paragraph \cite{kincaid1975derivation}, such that increases in text difficulty were associated with increased reading times. Crucially, their study yielded evidence of interactions between predictability (surprisal scores) and paragraph difficulty. In the above-mentioned computational study,  \citet{hollenstein-etal-2021-multilingual} also investigated the influence of textual characteristics (word length, text readability) on model performance. Text readability was measured using Flesch Reading Ease scores \cite{flesch1948new}. Their results indicated that the models learned to reflect characteristics of human reading, such as sensitivity to word length. They also found that model accuracy was higher in more easily readable sentences.

\section{Experiments}

\subsection{Datasets}

We analyze eye movement data from two eye-tracking corpora of natural reading, the Ghent Eye-Tracking Corpus (GECO; \cite{cop2017presenting}) and the Provo corpus \cite{luke2018provo}. In both corpora the participants read full sentences within longer spans of naturally occurring text at their own speed while their eye movements were recorded. The GECO corpus is large dataset of eye movement of a monolingual and bilingual readers who read a complete novel, Agatha Christie’s `The Mysterious Affair at Styles'. It contains eye-tracking data from 14 English native speakers and 19 bilingual speakers of Dutch and English, who read parts of the novel in its original English version and another part of its Dutch translation. In the present work, we focus on the analysis of the data from the monolingual English native speakers. These participants read a total of 5031 sentences amounting to a total of 54364 word tokens. The Provo Corpus is a dataset of eye movements of skilled readers reading connected text. It consists of eye movement data from 84 native English-speaking participants from Brigham Young University, who read 55 short passages from a variety of sources, including online news articles, popular science magazines, and public-domain works of fiction. These passages were an average of 50 words long for a total of 2,689 word tokens.

\subsection{Measurement of text properties}

The texts from both datasets (GECO and PROVO) were automatically analyzed using CoCoGen \cite{stroebel2016:cocogen}, a computational tool that implements a sliding window technique to calculate sentence-level measurements that capture the within-text distributions of scores for a given language feature (for current applications of the tool in the context of text classification, see \citet{kerz2020becoming, kerz-etal-2021-language}).  We extract a total of 107 features that fall into five categories:  (1) measures of syntactic complexity (N=16), (2) measures of lexical richness (N=14), (3) register-based n-gram frequency measures (N=25), (4) readability measures (N=14), and (5) psycholinguistic measures (N=38). A concise overview of the features used in this study is provided in Table \ref{tab:indicators} in the appendix. Tokenization, sentence splitting, part-of-speech tagging, lemmatization and syntactic PCFG parsing were performed using Stanford CoreNLP \cite{manning2014stanford}. The syntactic complexity measures comprise (i) surface measures that concern the length of production units, such as the mean length of words, clauses and sentences, (ii) measures of the type and incidence of embeddings, such as dependent clauses per T-Unit or verb phrases per sentence or (iii) the frequency of particular types of particular structures, such as the number of complex nominal per clause. These features are implemented based on descriptions in \citet{lu2010automatic} and using the Tregex tree pattern matching tool \cite{levy2006tregex} with syntactic parse trees for extracting specific patterns. Lexical richness measures fall into three distinct sub-types: (i) lexical density, such as the ratio of the number of lexical (as opposed to grammatical) words to the total number of words in a text, (iii) lexical variation, i.e. the range of vocabulary as displayed in language use, captured by text-size corrected type-token ratio and (iii) lexical sophistication, i.e. the proportion of relatively unusual or advanced words in the learner’s text, such as the number of New General Service List \cite{browne2013new}. The operationalizations of these measures follow those described in \citet{lu2012:relationship} and \citet{strobel2014:tracking}. The register-based n-gram frequency measures are derived from the five register sub-components of the Contemporary Corpus of American English (COCA, \cite{davies2008corpus}): spoken, magazine, fiction, news and academic language\footnote{The Contemporary Corpus of American English is the largest genre-balanced corpus of American English, which at the time the measures were derived comprised of 560 million words.}. These measures consider both the register-specific frequency rank and count:

%\scalebox{0.95}{
\begin{equation}
\text{Norm}_{n,s,r}=\frac{|C_{n,s,r}|\cdot \log\left[\prod_{c\in|C_{n,s,r}|}freq_{n,r}(c)\right]}{|U_{n,s}|}
\end{equation}
%}

 Let $A_{n,s}$ be the list of n-grams ($n\in[1, 5]$) appearing within a sentence $s$, $B_{n,r}$ the list of n-gram appearing in the n-gram frequency list of register $r$ ($r\in\{\text{acad}, \text{fic}, \text{mag}, \text{news}, \text{spok}\}$) and $C_{n,s,r}=A_{n,s} \cap B_{n,r}$ the list of n-grams appearing both in $s$ and the n-gram frequency list of register $r$. $U_{n,s}$ is defined as the list of unique n-gram in $s$, and $freq_{n,r}(a)$ the frequency of n-gram $a$ according to the n-gram frequency list of register $r$. The total of 25 measures results from the combination of (a) a `reference list' containing the top 100k most frequent n-grams and their frequencies from one of five registers of the COCA corpus and (b) the size of the n-gram ($n\in [1,5]$). The readability measures combine a word familiarity variable defined by prespecified vocabulary resource to estimate semantic difficulty together with a syntactic variable, such as average sentence length. Examples of these measures are the Fry index \cite{fry1968readability} or the SMOG \cite{mclaughlin1969clearing}. Finally, the psycholinguistic measures capture cognitive aspects of reading not directly addressed by the surface vocabulary and syntax features of traditional formulas. These measures include a word’s average age-of-acquisition \cite{kuperman2012age} or prevalence, which refers to the number of people knowing the word \cite{brysbaert2019word, johns2020estimating}. 
%Figure \ref{fig:examples} \textcolor{red}{presents examples of the extracted complexity contours for four  selected  measures of two randomly selected speeches.}  As is evident in the graphs, all features scores fluctuate within each speech and often display `compensatory' behavior, such that high scores in one feature are accompanied by low scores on another. 

\subsection{Eye-tracking measures}

We analyze data from eight word-level reading time measures, which were also investigated in \citet{hollenstein-etal-2021-multilingual}. The measures include general word-level characteristics such as (1) the number of fixations (NFX), i.e. the number of times a subject fixates on a given word w, averaged over all participants, (2) mean fixation duration (MFD), the average fixation duration of all fixations made on w, averaged over all participants and (3) fixation proportion (FXP), the number of subjects that fixated w, divided by the total number of participants. `Early processing' measures pertain to the early lexical and syntactic processing and are based on the first time a word is fixated. These features include: (4) first fixation duration (FFD), i.e. the duration of the first fixation on w (in milliseconds), averaged over all subjects and (5) first pass duration (FPD), i.e. the sum of all fixations on w from the first time a subject fixates w to the first time the subject fixates another token. `Late processing' measures capture the late syntactic processing and are based on words which were fixated more than once. These measures comprise (6) total fixation duration (TFD), i.e. the sum of the duration of all fixations made on w, averaged over all subjects, (7) number of re-fixations (NRFX), the number of times w is fixated after the first fixation, i.e., the maximum between 0 and the NFIX-1, averaged over all subjects and (8) re-read proportion (RRDP), the number of subjects that fixated w more than once, divided by the total number of subjects. The means, standard deviations and observed ranges for all eye-tracking features are shown in Tables \ref{tab:etGECO} and \ref{tab:etPROVO}. Like in \citet{hollenstein-etal-2021-multilingual}, before being entered into the models, all eye-tracking features were scaled between 0 and 100 so that the loss can be calculated uniformly over all features. 

\begin{table*}[h!]
	\setlength{\tabcolsep}{2pt}
	\centering
	\begin{minipage}[b]{0.4\textwidth}
		\begin{tabular}{lrrrr}
			\hline
			Feature & M & SD & Min & Max \\ 
			\hline
			NFX & 0.81 & 0.45 & 0.00 & 7.50 \\ 
			MFD & 128.41 & 58.98 & 0.00 & 350.92 \\
			FXP & 0.61 & 0.25 & 0.00 & 1.00 \\ 
			\hline 
			FFD & 129.28 & 60.06 & 0.00 & 371.31 \\ 
			FPD & 143.25 & 77.49 & 0.00 & 1425.86 \\ 
			\hline
			TFD & 168.20 & 102.44 & 0.00 & 1804.00 \\ 
			NRFX & 0.20 & 0.26 & 0.00 & 6.50 \\ 
			RRDP & 0.15 & 0.16 & 0.00 & 1.00 \\ 
			\hline
		\end{tabular}
		\caption{Descriptive statistics of eye-tracking measures for the GECO dataset.}
		\label{tab:etGECO}
	\end{minipage}\hspace{10mm}
\begin{minipage}[b]{.4\textwidth}
\begin{tabular}{lrrrr}
	\hline
	Feature & M & SD & Min & Max \\ 
	\hline
NFX & 0.95 & 0.47 & 0.13 & 3.61 \\ 
 MFD & 139.91 & 52.13 & 23.07 & 272.71 \\ 
  FXP & 0.66 & 0.22 & 0.13 & 1.00 \\
  \hline
   FFD & 139.83 & 52.02 & 23.18 & 276.86 \\ 
   FPD & 165.91 & 80.27 & 24.24 & 736.62 \\ 
   \hline
 TFD & 198.21 & 107.20 & 24.24 & 940.50 \\ 
 NRFX & 0.28 & 0.29 & 0.00 & 2.62 \\ 
 RRDP & 0.21 & 0.17 & 0.00 & 0.87 \\ 
	\hline
\end{tabular}
\caption{Descriptive statistics of eye-tracking measures for the PROVO dataset.}
\label{tab:etPROVO}
\end{minipage}
\end{table*}
%These features can be categorized into three groups: (1) Word level characteristics, (2) early processing features and (3) late processing features. Word level characteristics concern word-level fixations defined as the period of time where the gaze of a reader is maintained on a single location. 

\section{Modeling approach}

Deep neural transformer-based language models create contextualized word representations that are sensitive to the context in which the words appear. These models have yielded significant improvements on a diverse array of NLP tasks, ranging from question answering to coreference resolution. We compare two such models in terms of their ability to predict eye-tracking features: `Bidirectional Encoder Representations from Transformers' (BERT) \cite{devlin2018bert} and `Generative Pre-trained Transformer 2' (GPT-2) \cite{radford2019language}. BERT is an auto-encoder model trained with a dual objective function of predicting masked words and the next sentence. It consists of stacked transformer encoder blocks and uses self-attention, where each token in an input sentence looks at the bidirectional context, i.e. tokens on left and right of the considered token. In contrast, GPT-2 is an autoregressive model consisting of stacked transformer decoder blocks trained with a language modelling objective, where the given sequence of tokens is used to predict the next token. While GPT-2 uses self-attention as well, it employs masking to prevent words from attending to following tokens, hereby processing language fully unidirectionally. BERT is trained on the BooksCorpus (800M words) and Wikipedia (2,500M words), whereas GPT-2 is trained on WebText, an 8-million documents subset of CommonCrawl amounting to 40 GB of text. We chose the BERT base model (cased) because it is most comparable to GPT-2 with respect to number of layers and dimensionality (BERT base model (cased) has 110M trainable parameters, GPT-2 has 117M).

We evaluate the eye-tracking predictions of the models both on within-domain text, using an 80/10/10 split of the much larger GECO dataset (representing fiction language), as well as on out-of-domain text using the complete, much smaller PROVO dataset (comprising also online news and popular science magazine language). Furthermore, since overly aggressive fine-tuning may cause catastrophic forgetting \cite{howard2018universal}, we perform all experiments both with `frozen' language models, where all the layers of the language model are frozen and only the attached neural network layers are trained, and also `fully fine-tuned' language models, where the error is back-propagated through the entire architecture and the pretrained weights of the model are updated based on the GECO training set. 

For all models we explored in this paper, we apply a dropout rate of 0.1 and a l2 regularization of $1\times10^{-4}$. We use AdamW as the optimizer and mean squared error as the loss function. We use a fixed learning rate with warmup. During warmup, the learning rates are linearly increased to the peak learning rates and then fixed. For BERT with a `frozen' language model, the peak learning rate is $5\times10^{-4}$ with 5 warmup steps and for GPT-2 with a `frozen' language model, it is $0.001$ also with 5 warmup steps. Models with 'fully fine-tuned' language models are trained with two phases. In the first phase, the weights of the language models are frozen and only regression layers are trained. During this phase, peak learning rates of $3\times10^{-4}$ for BERT and $0.001$ for GPT-2 are used. For both models, the first phase is performed over 12 epochs with 5 warmup steps. In the second phase, we unfreeze the weights of language models and fine-tune the language models together with the regression layers. During this phase, the BERT-based model is trained with a peak learning rate of $5\times10^{-5}$ while GPT-2-based model is trained with a peak learning rate of $5\times10^{-4}$. The number of warmup steps for training both models in this phase is 3. We adopted a two-phase training procedure since preliminary experiments showed that this procedure yields same results as training the entire models from the first epoch, yet it can speed up model convergence. All hyper-parameters are optimized through grid search.

% Feel free to use/adapt form this
% \begin{itemize}
%     \item The eye-tracking prediction uses a model for token regression, i.e., the pretrained and fine-tuned language models with a linear dense layer on top of it. We use the pretrained checkpoints \textit{bert-base-uncased} and \textit{GPT-2} from the HuggingFace library to build our regression models. Concretely, we evaluate the pretrained checkpoints only and furthermore measure both models' performances when performing task-specific fine-tuning on all of their 12 layers. 
% \end{itemize}

\subsection{Influence of text characteristics on model performance}

To investigate the impact of the text properties listed in Section 3.2 on prediction accuracy, we partitioned the GECO testset into deciles according to each textual property, i.e. each of the 107 features. We then calculated the Pearson correlation coefficients between the decile of a given textual feature and the mean absolute error (MAE) of a given model. We expected to observe higher prediction accuracy (lower MAE) for sentences with higher readability, lower syntactic complexity, lower lexical richness, higher n-gram frequency and less demanding psycholinguistic properties, i.e. lower age-of-acquisition scores and higher prevalence scores.  

\subsection{Integration of text characteristics using a hybrid modeling approach}

\begin{figure}
    \centering
    \includegraphics[width = 0.48\textwidth]{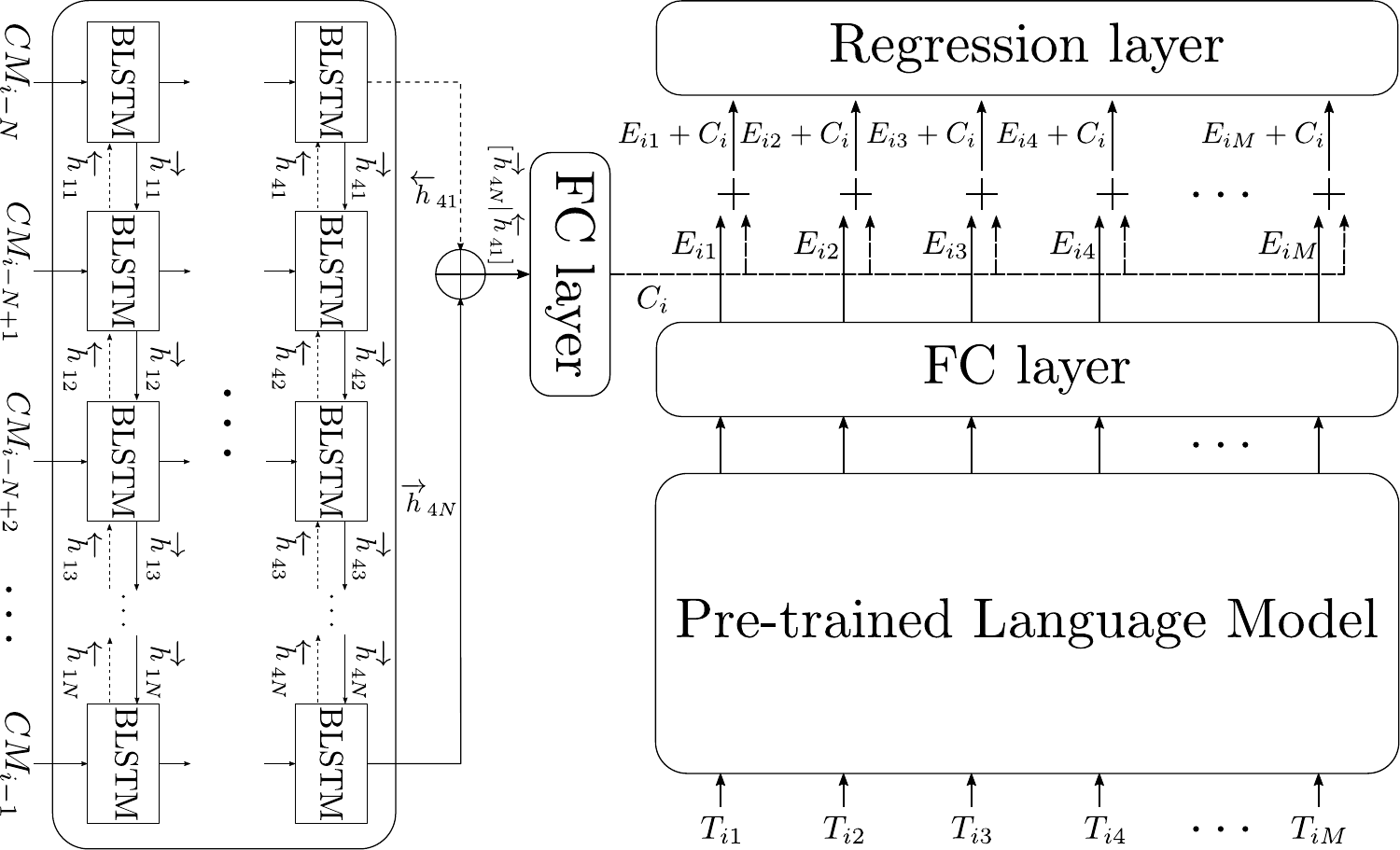}
    \caption{Visualization of approach used to integrate information on complexity of preceding language input for sentence $i$.}
    \label{fig:ctx_lm}
    \vspace{-5mm}
\end{figure}

%\textcolor{purple}{\textbf{@Yu:} HERE PLEASE PROVIDE A DESCRIPTION OF EXTENDED MODEL ARCHITECTURE}  
To determine whether eye movement patterns were affected by textual characteristics of the previous sentences (sentence spillover effects), a bidirectional LSTM (BLSTM) model was integrated into the predictive models (Figure \ref{fig:ctx_lm}). This BLSTM model reads 107 dimensional vectors of textual features $CM_{i-N}, \cdots, CM_{i-1}$ from $N$ previous sentences\footnote{Experiments with $N\in[1,5]$ were performed and $N=1$ performed best.} as its input, transforms them through 4 BLSTM layers of 512 hidden units each, and outputs a 1024 dimensional vector $[\overrightarrow{h}_{4N}|\overleftarrow{h}_{41}]$, that is a concatenation of the last hidden states of the 4th BLSTM layer in the forward and backward directions $\overrightarrow{h}_{4N}, \overleftarrow{h}_{41}$. A fully connected (FC) layer is added on top of the BLSTM layers to reduce the dimension of BLSTM model output to 256 ($C_i$). Meanwhile, another FC layer is added to the pre-trained language model (BERT or GPT-2) in order to reduce its logits to the same dimension ($E_{i1},\cdots,E_{iM}$). The reduced BLSTM output is then added to each of the reduced language model logits. Finally, the 256-dimensional joint vectors are fed to a final regression layer to predict human reading behavior. The procedures used to train the `hybrid' models with textual characteristics of the previous sentences was identical to those specified above. Grid search yielded the same optimized values for all hyper-parameters, except for the peak learning rate of `fully fine-tuned' model with GPT-2 in second training phase, which was $1\times10^{-4}$.

To assess the relative importance of the feature groups, we employed Submodular Pick Lime (SP-LIME; \citet{ribeiro2016should}), a method to construct a global explanation of a model by aggregating the weights of the linear models. %The linear models serve as approximations of a complex model around small regions on the data manifold. To this end
We first construct local explanations using LIME with a linear local explanatory model, exponential kernel function with Hamming distance and a kernel width of $\sigma=0.75\sqrt{d}$, where $d$ is the number of feature groups. The global importance score of the SP-LIME for a given feature group $j$ can then be derived by: $I_j = \sqrt{\sum_{i=1}^n |W_{ij}|}$, where $W_{ij}$ is the $j$th coefficient of the fitted linear regression model to explain a data sample $x_i$.
% We first constructed local explanations using LIME. Analogous to super-pixels for images, we categorized our features into 5 (psycho-) groups and used binary vectors $z\in\{0,1\}^{d}$ to denote the absence and presence of feature groups in the perturbed data samples, where $d$ is the number of feature groups. Here, absent means that all values of the features in the feature group are set to 0, and present means that their values are retained. For simplicity, a linear regression model was chosen as the local explanatory model. An exponential kernel function with Hamming distance and kernel width $\sigma=0.75\sqrt{d}$ was used to assign different weights to each perturbed data sample. After constructing their local explanation for each data sample in the original dataset, the matrix $W\in\mathbb{R}^{n\times d}$ was obtained, where $n$ is the number of data samples in the original dataset and $W_{ij}$ is the $j$th coefficient of the fitted linear regression model to explain data sample $x_i$. The global importance score of the SP-LIME for feature $j$ can then be derived by: $I_j = \sqrt{\sum_{i=1}^n |W_{ij}|}$

%As models without considering previous sentences, we compared models with 'frozen' and 'fully fine-tuned' language models. Same training procedures were applied to train all models with textual characteristics of the previous sentences. Grid search yields the same optimized value for hyper-parameters expect for the peak learning rate of 'fully fine-tuned' model with GPT-2 language model in second training phase, which is $1\times10^{-4}$.

\FloatBarrier

\section{Results \& Discussion}

We use sentence-level accuracy (100-MAE) and coefficients of determination ($R2$) as metrics to evaluate the performance of all models. Table \ref{tab:fullresults} shows the evaluation results for all models averaged over all eye-tracking features. Table \ref{tab:fullresults} shows that both BERT and GPT-2 models predicted the eye-tracking features of both datasets with more than 92\% accuracy. The fine-tuned models performed consistently better than the pretrained-only (`frozen') models both on the within-domain text (GECO) and on the out-of-domain text (PROVO). This result indicates that the learned representations are general enough to be successfully applied both in the prediction of reading patterns of fiction texts as well as in the prediction of news and popular science texts. The BERT models consistently outperformed the GPT-2 models with a difference in $R2$ of as much as 10.54\% on the within-domain data (GECO). This result stands in sharp contrast with those reported in \citet{schrimpf2020neural} summarised in Section 2. In their interpretation of the success of GPT-2 in predicting neural activity during reading, \citet{schrimpf2020neural} state that ``GPT-2 is also arguably the most cognitively plausible of the transformer models (because it uses unidirectional, forward attention)''. Especially in view of the remarkable margin by which the BERT models outperformed the GPT-2 models here, it appears that arguments that infer cognitive plausibility from prediction success should be viewed with caution (see also \citet{merkx2020human} for further intricacies of the issue). The most accurately predicted individual eye-tracking measures were fixation probability (FXP), mean fixation duration (MFD) and first fixation duration (FFD), indicating that prediction accuracy was generally better for early measures than for late measures. A detailed overview of the results for each eye-tracking measure across all models and datasets is provided in Table \ref{tab:detailed} in the appendix. This finding suggests that the accurate prediction of late measures – that are assumed to reflect higher order processes such as syntactic and semantic integration, revision, and ambiguity resolution – may benefit from the inclusion of contextual information beyond the current sentence.

\begin{table}
    \centering
        \caption{Model performance across datasets.}
        \vspace{-2mm}
        %\small
         \setlength{\tabcolsep}{3pt}
    \begin{tabular}{|l|l|c|c|c|}
\hline
Model&Dataset&R2(\%)&MAE&Acc\\
\hline
\multirow{2}{*}{\shortstack[l]{BERT fr  }}&GECO&42.14&7.01&92.99\\\cline{2-5}
&PROVO&42.19&6.93&93.61\\
\hline
\multirow{2}{*}{\shortstack[l]{BERT fr\\+ com S-1}}&GECO&43.29&6.93&93.07\\\cline{2-5}
&PROVO&51.70&5.74&94.26\\
\hline
\multirow{2}{*}{\shortstack[l]{BERT ft }}&GECO&56.83&5.95&94.05\\\cline{2-5}
&PROVO&67.64&4.51&95.49\\
\hline
\multirow{2}{*}{\shortstack[l]{BERT ft \\ + com S-1}}&GECO&\textbf{58.36}&\textbf{5.92}&\textbf{94.08}\\\cline{2-5}
&PROVO &\textbf{68.59}&\textbf{4.49}&\textbf{95.51}\\
\hline
\hline
\multirow{2}{*}{\shortstack[l]{GPT-2 fr  }}&GECO&35.00&7.32&92.68\\\cline{2-5}
&PROVO&40.15&6.26&93.74\\
\hline
\multirow{2}{*}{\shortstack[l]{GPT-2 fr \\ + com S-1}}&GECO &35.19&7.32&92.68\\\cline{2-5}
&PROVO&43.67&6.08&93.92\\
\hline
\multirow{2}{*}{\shortstack[l]{GPT-2 ft  }}&GECO&46.29&6.48&93.52\\\cline{2-5}
&PROVO&55.73&5.06&94.94\\
\hline
\multirow{2}{*}{\shortstack[l]{GPT-2 ft \\ + com S-1}}&GECO&47.53&6.38&93.62\\\cline{2-5}
&PROVO&56.77&5.08&94.92\\
\hline
\end{tabular}
\vspace{1ex}

     {\raggedright \textbf{Note:} `fr' =  freeze all layers of language model; `ft' = the entire model is fine-tuned; `+ com S-1' = including textual features of previous sentence \par}

    \label{tab:fullresults}
    % \vspace{-3mm}
\end{table}

\subsection{Relationship of prediction accuracy and text characteristics}

The correlation analyses of the textual features and the mean absolute error revealed that prediction accuracy was affected by the text characteristics of the sentence under consideration. Such effects were found across all eye-tracking metrics for both BERT and GPT-2 models in both their frozen and fully fine-tuned variants. For reasons of space, we focus our discussion on the predictions of the BERT frozen model of first pass durations on the GECO dataset  (additional results for both frozen and fine-tuned BERT models for both first pass duration and total fixation duration are provided in Figure \ref{fig:moreImp} in the appendix). Figure \ref{fig:BERTfpd} visualizes the impact of all textual features that reached correlation coefficients $r > |0.2|$ along with the feature group they belong to. As is evident in Figure \ref{fig:BERTfpd} the prediction accuracy of the BERT frozen model was impacted by features from all five feature groups with individual features affecting prediction accuracy in opposite ways. A strong impact ($r > |0.5|$) was observed for several features of the n-gram feature group: Fixation durations of sentences with higher scores on ngram-frequency features from the news, magazine and spoken registers were predicted more accurately than those with lower scores on these measures. The SMOG readability index, which estimates the years of education a person needs to understand a piece of writing, also has a strong impact: Predicted first pass durations were less accurate in sentences with higher SMOG scores.  Several features from the lexical richness, syntactic complexity and readability groups had a moderate impact on prediction accuracy ($|0.3| < r < |0.5|$):  For example, predictions of fixation durations were less accurate on sentences of with a more clausal embedding (ClausesPerSentence) and greater lexical sophistication (MeanLengthWord, Sophistication.ANC and Sophistication.BNC). A similar effect was also observed for the psycholinguistic age-of-acquisition features (AoA mean, AoA max), where predictions of fixations times were less accurate for later acquired words. Note that the finding that the correlation coefficients  of the readability features have opposite signs is due to the fact that these are either defined to quantify ease of reading (e.g. Flesch Kincaid Reading Ease) or reading difficulty (e.g. SMOG index). 

\begin{figure}
    \centering
     \includegraphics[width=0.9\linewidth]{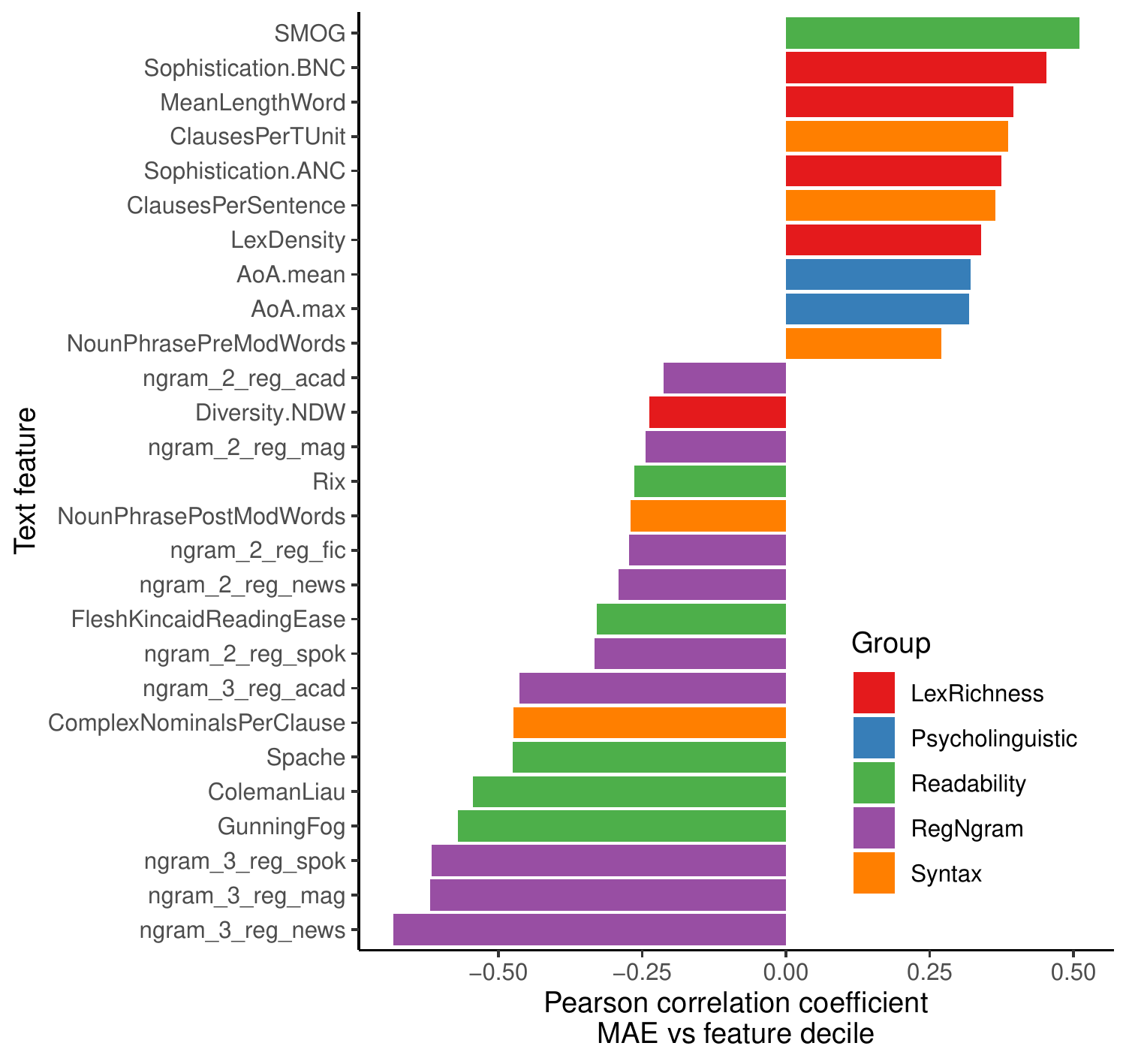} 
    %  \vspace{-2mm}
        \caption{Pearson correlations between model performance (mean absolute error), and the deciles of the respective text characteristics. For measures with negative correlation coefficients, model performance increased with higher values of the text characteristics (Data from `BERT frozen' predictions of First pass duration (FPD) on GECO testdata).} 
        \label{fig:BERTfpd}
        % \vspace{-4mm}
\end{figure}

\subsection{Prediction accuracy of hybrid models}

Turning to the results of the hybrid models with integrated information on textual characteristics of the preceding sentence, we found that highest accuracy ($R2 = 58.36$\%) was achieved by the fine-tuned BERT model. This amounts to an increase in performance over a model trained without that information of 1.53\%. This result demonstrates that future studies should take textual spillover effects into account. Our best-fitting model outperformed not only the best-performing BERT model in \citet{hollenstein-etal-2021-multilingual}, BERT-BASE-MULTILINGUAL-CASED \cite{wolf2019huggingface} but also the overall best-performing transformer-based model, XLM-MLM-ENDE-1024 \cite{lample2019cross} tested in that study. This result demonstrates that the claim put forth in \citet{hollenstein-etal-2021-multilingual} that multilingual models show an advantage over language specific ones and that multilingual models might provide cognitively more plausible representations in predicting reading needs to be viewed with caution.

The results of the feature ablation experiments revealed that the main sources of the greater prediction accuracy of the hybrid models was associated with information concerning the syntactic complexity, lexical richness and n-gram frequency of the preceding sentence. An overview of the results is presented in Table \ref{tab:ablation}. We focus here on the results on the out-of-domain testset (PROVO) for which improvements over models without the integrated textual information were more pronounced. As is evident in Table \ref{tab:ablation}, the central role of the three feature groups listed above result was observed across models (BERT vs. GPT-2) and across training procedures (frozen vs. fine-tuning). However, Table \ref{tab:ablation} also demonstrates clear differences between the models: While the BERT models show greater sensitivity to syntactic complexity, the GPT-2 models mostly benefit from information concerning n-gram frequency. A possible interpretation  of this finding is that a unidirectional model like GPT-2 relies more strongly on word sequencing than a bidirectional one. Future research is needed to examine this in more detail so that effects associated with differences in model architecture can be disentangled.

\begin{table}
\centering
 \setlength{\tabcolsep}{3pt}
\caption{Feature ablation of different models on PROVO dataset. Most important feature groups are bolded.}
\vspace{-2mm}
% \begin{tabular}{|lr|lr|}
% \hline
% \multicolumn{2}{|c}{BERT frozen}&\multicolumn{2}{|c|}{BERT ft}\\
% group&score&group&score\\
% \hline
% \hline
% Syn complex &5.69&Syn complex&2.48\\
% Reg. n-gram&5.44&Lex. richness&1.64\\
% Lex. richness&5.34&Reg. n-gram&1.44\\
% Readability&4.74&Readability&1.30\\
% Psycholing&3.22&Psycholing&1.25\\
% \hline
% \hline
% \multicolumn{2}{|c}{GPT-2}&\multicolumn{2}{|c|}{GPT-2\_ft}\\
% group&score&group&score\\
% \hline
% \hline
% Reg. n-gram&10.10&Reg. n-gram&18.82\\
% Readability&9.85&Syn complex &16.07\\
% Lex. richness&9.62&Readability&12.68\\
% Syn complex &9.06&Psycholing&12.20\\
% Psycholing&7.91&Lex. richness&7.02\\
% \hline
% \end{tabular}

% \setlength{\tabcolsep}{3pt}
\begin{tabular}{|cl|ccccc|}
		\hline
		&&  \multirow{2}{*}{\shortstack[c]{Syn.\\complex}} &  \multirow{2}{*}{\shortstack[c]{Lex.\\richness}}&  Psych. &  \multirow{2}{*}{\shortstack[c]{Reg.\\ngram}} &  Read. \\
		&&&&&&\\
		\hline
		&&&&&&\\[\dimexpr-\normalbaselineskip+2.5pt]
		\multirow{2}{*}{\rotatebox{90}{BERT}} &fr &         \textbf{5.69} &           5.34 &        3.22 &         5.44 &         4.74 \\
		 &ft     &          \textbf{2.48} &           1.64 &        1.25 &         1.44 &         1.30 \\
		 &&&&&&\\[\dimexpr-\normalbaselineskip+2pt]
		 \hline
		 \hline
		 &&&&&&\\[\dimexpr-\normalbaselineskip+2pt]
			\multirow{2}{*}{\rotatebox{90}{GPT-2}}  &fr &         9.06 &           9.62 &        7.91 &         \textbf{10.10} &         9.85 \\
		&ft     &        16.07 &           7.02 &       12.20 &         \textbf{18.82} &        12.68 \\[1.5pt]
%		&&&&&&\\[\dimexpr-\normalbaselineskip+2pt]
		\hline
	\end{tabular}

\label{tab:ablation}
\vspace{-3mm}
\end{table}

\section{Conclusion}

%The particular utility of eye movement data has been recognized in several areas of natural language processing and cognitive science. 
%Eye-tracking data ... can not only be employed to improve the performance of machine‐learning models \cite{bakarov2018can}, they can also reveal the workings of the underlying cognitive processes of human language processing. Transformer-based language models have proven to be impressive in their ability to predict human reading patterns and promise to be a valuable tool for the development mechanistic accounts of human language processing in general and reading in particular. 

%n this paper, we investigated the role of textual properties, assessed through a state-of-the-art automatic text analysis system, in the prediction of human reading behavior using transformer-based language models (BERT \& GPT-2). We demonstrated (1) that model accuracy is systematically linked to sentence-level textual features spanning across five measurement categories (syntactic, complexity, lexical richness, register-specific n-gram frequency, readablity and psycholinguistic properties) and showed (2) that prediction accuracy can be improved through the use of hybrid models that take into account spillover effects from the previous sentence.

 In this paper we conducted the first systematic investigation of the role of general text features in predicting human reading behavior using transformer-based language models (BERT \& GPT-2). We have shown (1) that model accuracy is systematically linked to sentence-level text features spanning five measurement categories (syntax, complexity, lexical richness, register-specific N-gram frequency, readability, and psycholinguistic properties), and (2) that prediction accuracy can be improved by using hybrid models that consider spillover effects from the previous sentence.

\bibliography{anthology,custom}
\bibliographystyle{acl_natbib}

\newpage
\FloatBarrier

\appendix
\onecolumn
\section{Appendix}
\label{sec:appendix}

\begin{table*}[h!]
  \centering
  \setlength{\tabcolsep}{2pt}
  \caption{Overview of the 107 features investigated in the work}
    \begin{tabular}{|l|c|l|l|}
		\hline
		Feature group & Number & Features & Example/Description \\
		& of features &  & \\
		\hline
		Syntactic complexity & 16    &MLC&Mean length of clause (words)\\

		&&MLS&Mean length of sentence (words)\\
		&&MLT&Mean length of T-unit (words)\\
		&&C/S&Clauses per sentence\\
		&&C/T&Clauses per T-unit\\
		&&DepC/C&Dependent clauses per clause\\
		&&T/S&T-units per sentence\\
		&&CompT/T&Complex T-unit per T-unit\\
		&&DepC/T&Dependent Clause per T-unit\\
		&&CoordP/C&Coordinate phrases per clause\\
		&&CoordP/T&Coordinate phrases per T-unit\\
		&&NP.PostMod&NP post-mod (word)\\
		&&NP.PreMod&NP pre-mod (word)\\
		&&CompN/C&Complex nominals per clause\\
		&&CompN/T&Complex nominals per T-unit\\
		&&VP/T&Verb phrases per T-unit\\
		\hline
		Lexical richness & 14    &MLWc& Mean length per word  (characters)\\
		&&MLWs&Mean length per word (sylables)\\
		&&LD&Lexical density\\
		&&NDW&Number of different words\\
		&&CNDW&NDW corrected by Number of words\\
		&&TTR&Type-Token Ration (TTR)\\
		&&cTTR&Corrected TTR\\
		&&rTTR&Root TTR\\
		&&AFL&Sequences Academic Formula List\\
		&&ANC& LS (ANC) (top 2000, inverted)\\
		&&BNC&LS (BNC) (top 2000, inverted)\\
		&&NAWL&LS New Academic Word List\\
		&&NGSL&LS (General Service List) (inverted)\\
		&&NonStopWordsRate& Ratio of words in NLTK non-stopword list\\
		\hline
		Register-based  & 25    & Spoken ($n\in [1,5]$) & Frequencies of uni-, bi-\\
		&       & Fiction ($n\in [1,5]$) & tri-, four-, five-grams\\
		&       & Magazine ($n\in [1,5]$) &  from the five sub-components \\
		&       & News ($n\in [1,5]$) & (genres) of the COCA,\\
		&       & Academic ($n\in [1,5]$) &  see \citet{davies2008corpus}\\
		\hline
	\end{tabular}%
  \label{tab:indicators}%
\end{table*}

\begin{table*}
  \centering
  \setlength{\tabcolsep}{2pt}
  \caption{Overview of the 107 features investigated in the work(Cont.}
    \begin{tabular}{|l|c|l|l|}
		\hline
		Feature group & Number & Features & Example/Description \\
		& of features &  & \\
		\hline
		Readability & 14     & ARI  &  Automated Readability Index \\
		&       & ColemanLiau & Coleman-Liau Index \\
		&       &DaleChall &  Dale-Chall readability score\\
		&&FleshKincaidGradeLevel&Flesch-Kincaid Grade Level\\ 
		&&FleshKincaidReadingEase&Flesch Reading Ease score\\
		&&Fry-x&x coord. on Fry Readability Graph\\
		&&Fry-y&y coord. on Fry Readability Graph\\
		&&Lix&Lix readability score\\
		&&SMOG&Simple Measure of Gobbledygook\\
		&&GunningFog&Gunning Fog Index readability score\\
		&&DaleChallPSK&Powers-Sumner-Kearl Variation of \\
		&&&the Dale and Chall Readability score\\
		&&FORCAST&FORCAST readability score\\
		&&Rix&Rix readability score\\
		&&Spache&Spache readability score\\
		
		\hline
		Psycholinguistic  & 38    &  WordPrevalence & See \citet{brysbaert2019word} \\
		&& Prevalence & Word prevalence list\\
		&&&incl. 35 categories (\citet{johns2020estimating})\\
		&&AoA-mean&avg. age of acquisition (\citet{kuperman2012age})\\
		&&AoA-max&max. age of acquisition\\
		\hline
	\end{tabular}%
\end{table*}

\begin{comment}
\begin{figure}
    \centering
    \caption{Mean scores of the eight eye-tracking features for both eye-tracking corpora}
    \includegraphics[width=0.5\textwidth]{figures/meanETperformanceComp.png}
    \label{fig:etCORPUS}
\end{figure}

\begin{figure}
    \centering
    \caption{Inter-correlations among the eight eye-tracking measures.}
    \includegraphics[width = 0.5\textwidth]{figures/correlationsET.png}
    \label{fig:my_label}
\end{figure}
\end{comment}

\begin{table*}
    \centering
    \caption{Model performance by eye-tracking feature across datasets}
    \begin{tabular}{|l|l|c|c|c|c|c|c|c|c||c|}
\hline
model&dataset&\multicolumn{8}{c||}{R2(\%)}&mean\\
&&NFX&FFD&FPD&TFD&MFD&FXP&NRFX&RRDP&R2(\%)\\
\hline
\hline
\multirow{3}{*}{BERT frozen}&GECO dev&46.38&44.50&46.22&45.21&44.62&46.01&31.80&34.50&42.40\\\cline{2-11}
&GECO test&46.99&42.60&45.34&45.05&42.94&44.91&33.68&35.61&42.14\\\cline{2-11}
&PROVO&42.91&50.28&46.11&42.83&48.99&44.76&29.67&31.95&42.19\\
\hline
\multirow{3}{*}{\shortstack[l]{BERT frozen \\ + complexity S-1}}&GECO dev&46.99&46.31&46.81&45.81&46.42&48.65&31.79&35.05&43.48\\\cline{2-11}
&GECO test&47.76&44.36&45.96&45.78&44.80&47.80&33.79&36.09&43.29\\\cline{2-11}
&PROVO&52.20&61.44&54.03&50.12&60.15&61.53&33.96&40.19&51.70\\
\hline
\hline
\multirow{3}{*}{BERT fine-tuned}&GECO dev&60.89&56.98&58.64&59.15&57.15&60.60&47.60&51.11&56.51\\\cline{2-11}
&GECO test&61.67&56.47&58.28&59.74&57.10&60.62&49.09&51.67&56.83\\\cline{2-11}
&PROVO&68.81&74.80&68.20&65.86&74.93&78.06&53.01&57.46&67.64\\
\hline
\multirow{3}{*}{\shortstack[l]{BERT fine-tuned \\ + complexity S-1}}&GECO dev&62.50&57.89&60.85&60.47&58.09&61.14&47.61&50.03&57.32\\\cline{2-11}
&GECO test&\textbf{64.17}&\textbf{57.66}&\textbf{61.59}&\textbf{61.83}&\textbf{58.20}&\textbf{61.27}&\textbf{50.14}&\textbf{52.00}&\textbf{58.36}\\\cline{2-11}
&PROVO&\textbf{70.49}&\textbf{75.39}&\textbf{70.05}&\textbf{67.16}&\textbf{75.21}&\textbf{77.60}&\textbf{52.54}&\textbf{60.27}&\textbf{68.59}\\
\hline

\hline
\hline
\multirow{3}{*}{GPT-2 frozen}&GECO dev&41.06&40.69&41.26&39.54&40.83&42.73&25.94&29.30&37.67\\\cline{2-11}
&GECO test&38.01&38.08&38.69&36.08&38.19&40.55&23.43&26.98&35.00\\\cline{2-11}
&PROVO&38.40&51.14&43.17&38.36&50.06&47.19&23.43&29.41&40.15\\
\hline
\multirow{3}{*}{\shortstack[l]{GPT-2 frozen  \\ + complexity S-1}}&GECO dev&41.02&41.14&41.16&39.46&41.27&43.97&25.27&29.28&37.82\\\cline{2-11}
&GECO test&37.98&38.55&38.56&36.08&38.66&41.81&22.70&27.19&35.19\\\cline{2-11}
&PROVO&43.09&54.07&45.78&41.77&52.65&53.37&27.49&31.14&43.67\\
\hline
\multirow{3}{*}{GPT-2 fine-tuned}&GECO dev&52.17&51.63&51.83&49.97&51.79&55.68&33.36&37.70&48.02\\\cline{2-11}
&GECO test&50.65&49.69&49.71&47.86&49.97&54.13&32.24&36.09&46.29\\\cline{2-11}
&PROVO&55.02&67.48&56.59&52.43&66.64&68.82&35.56&43.27&55.73\\
\hline
\multirow{3}{*}{\shortstack[l]{GPT-2 fine-tuned  \\ + complexity S-1}}&GECO dev&54.46&53.34&53.91&52.21&53.69&57.20&35.11&39.63&49.94\\\cline{2-11}
&GECO test&\textbf{51.91}&\textbf{50.97}&\textbf{51.24}&\textbf{49.30}&\textbf{51.28}&\textbf{55.02}&\textbf{33.11}&\textbf{37.42}&\textbf{47.53}\\\cline{2-11}
&PROVO&\textbf{56.19}&\textbf{68.20}&\textbf{58.44}&\textbf{53.98}&\textbf{67.84}&\textbf{68.79}&\textbf{35.81}&\textbf{44.95}&\textbf{56.77}\\
\hline

\end{tabular}
    \vspace{1ex}

     {\raggedright \textbf{Note:} `frozen' =  all the layers of the language model are frozen and only the attached neural network layers are trained on the GECO dataset; the weights of only the attached layers will be updated during model training. `fine-tuned' = the entire pretrained model is fine-tuned on the GECO training set; the error is back-propagated through the entire architecture and the pre-trained weights of the model are updated based on the GECO training set. Best-performing models on the two testsets (GECO test, PROVO) are highlighted in bold. \par}
     
    \label{tab:detailed}
\end{table*}

\begin{figure*}
    \centering
    
    %% (a)
    \begin{subfigure}[t]{0.45\textwidth}
        \centering
        \includegraphics[width=1\linewidth]{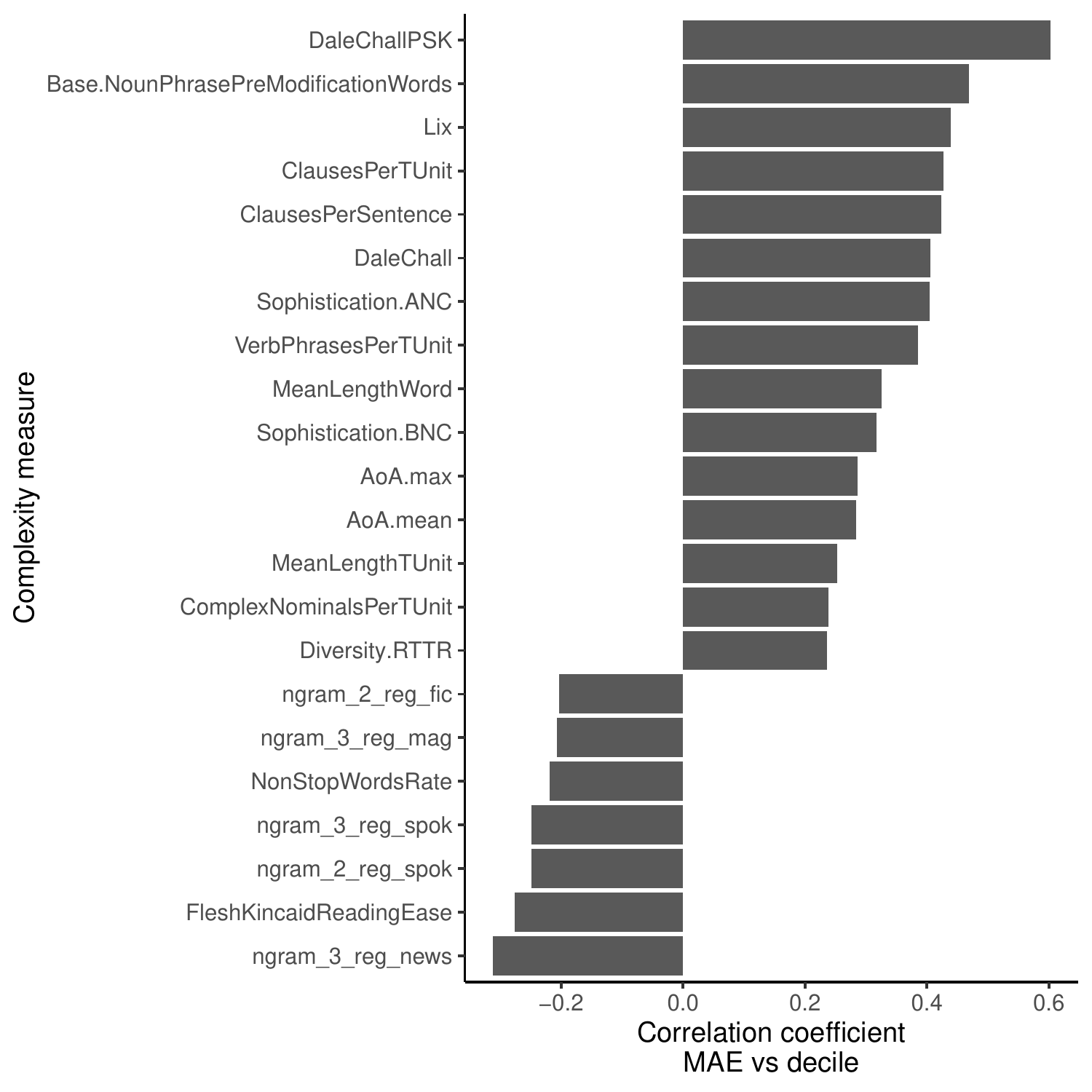} 
        \caption{Model: BERT frozen. Eye-tracking metric:  Total fixation duration (TFD)} \label{fig:timing1}
    \end{subfigure}
    \hfill
        %% (b)
    \begin{subfigure}[t]{0.45\textwidth}
        \centering
        \includegraphics[width=0.9\linewidth]{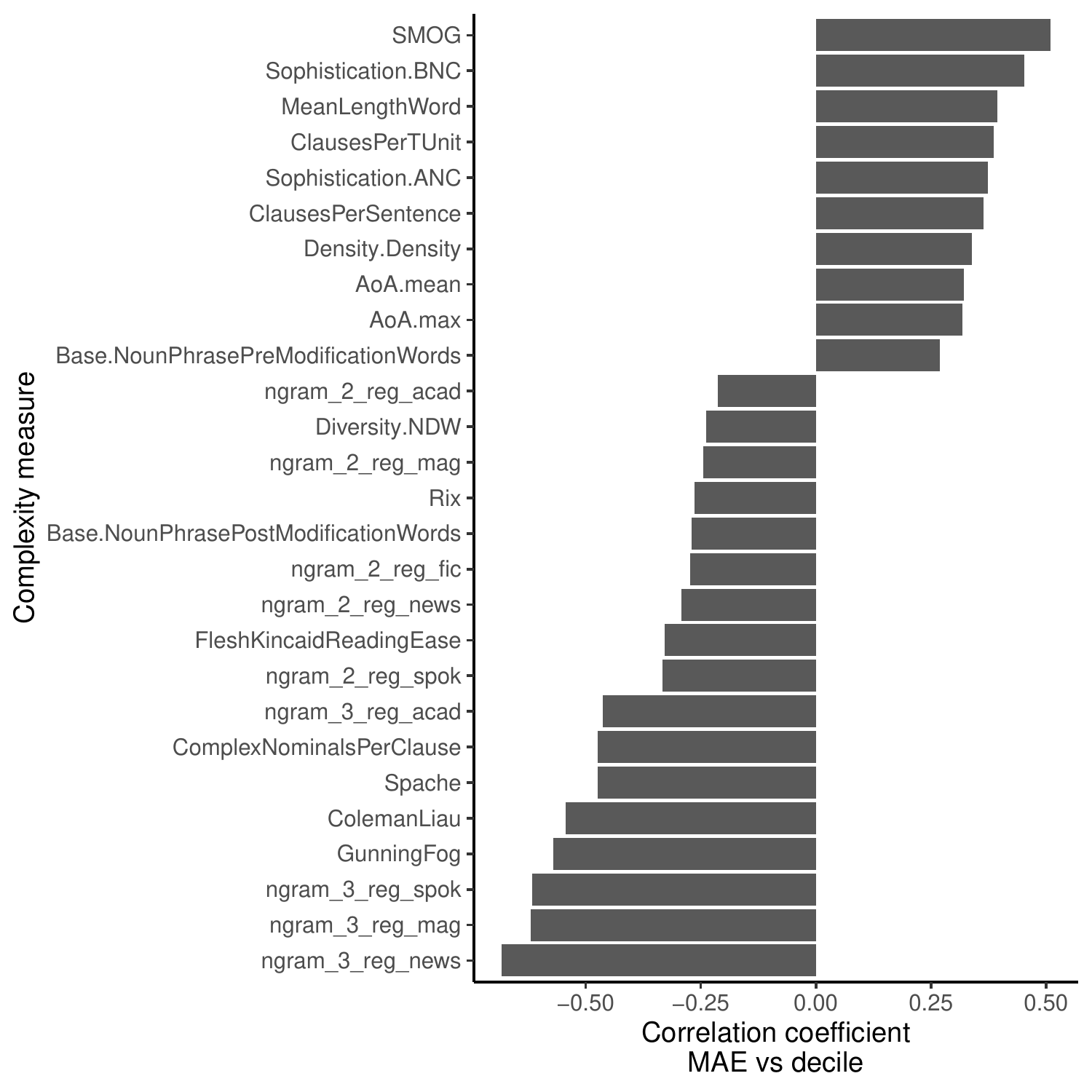} 
        \caption{Model: BERT frozen. Eye-tracking metric:  First pass duration (FPD)} \label{fig:timing2}
    \end{subfigure}

    \vspace{1cm}
        %% (c)
    \begin{subfigure}[t]{0.45\textwidth}
        \centering
        \includegraphics[width=1\linewidth]{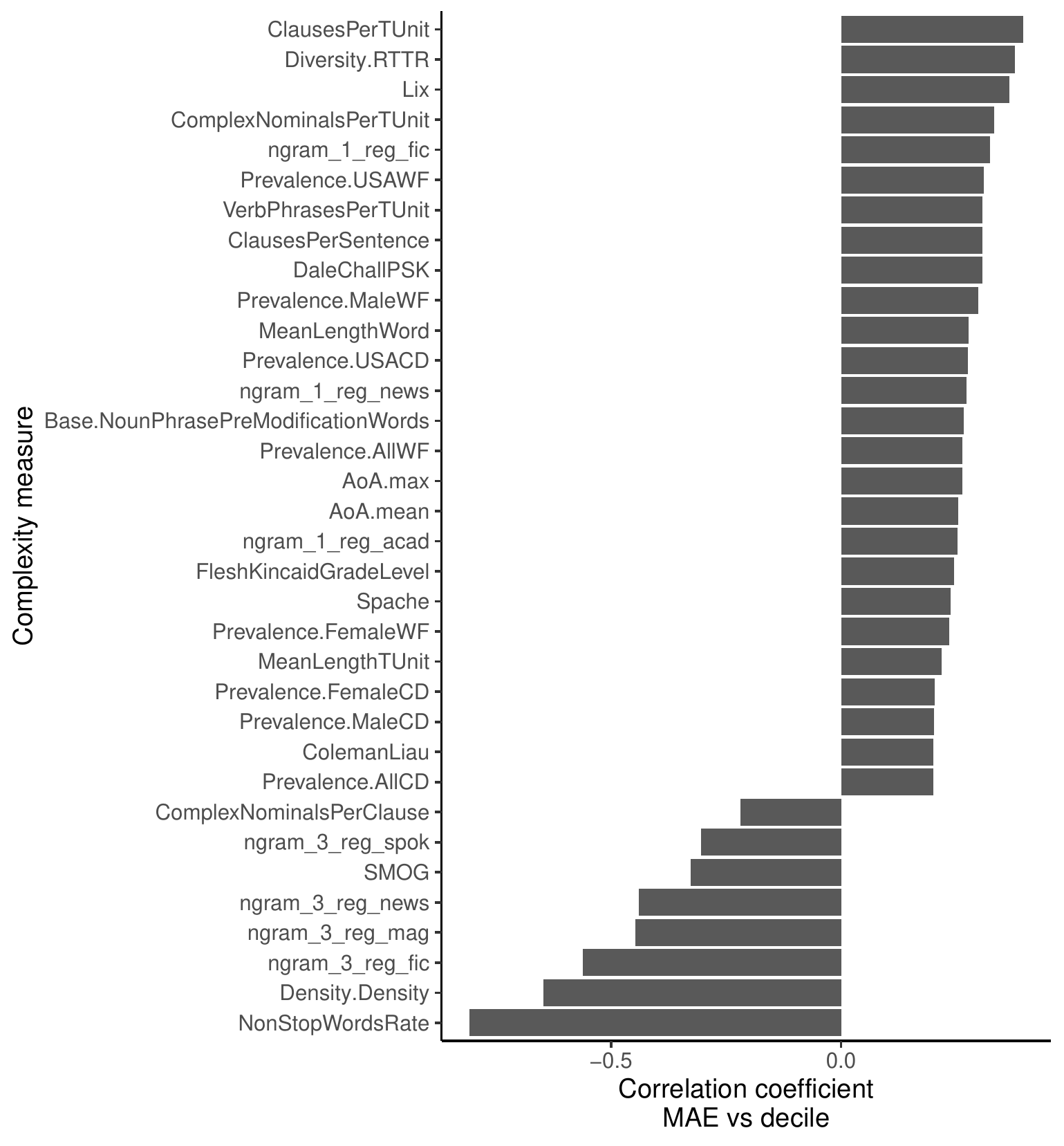} 
        \caption{Model: BERT fine-tuned. Eye-tracking metric:  Total fixation duration (TFD)} \label{fig:timing2}
        \label{fig:timing3}
    \end{subfigure}
    \hfill
        %% (d)
    \begin{subfigure}[t]{0.45\textwidth}
    \includegraphics[width=1\linewidth]{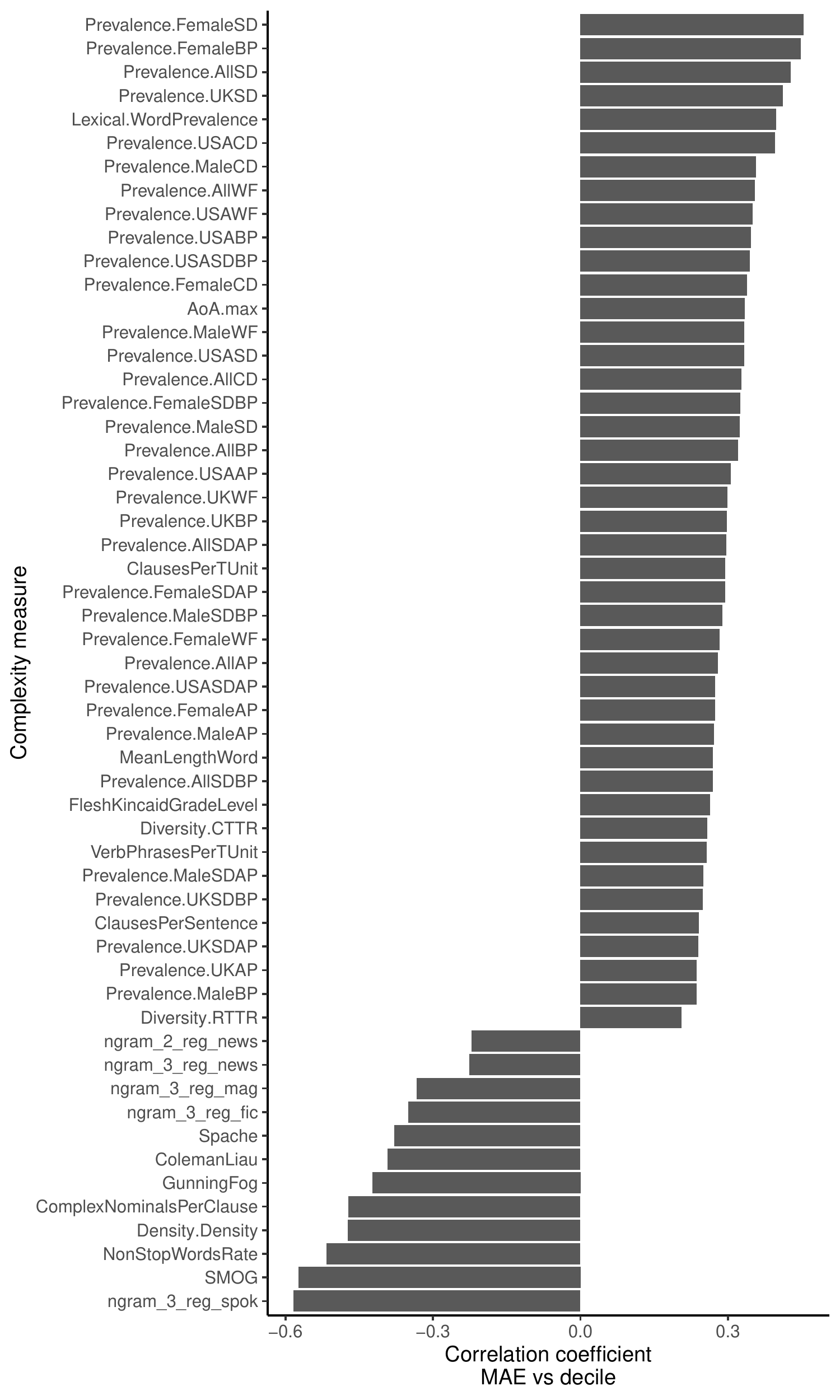} 
        \caption{Model: BERT fine-tuned. Eye-tracking metric:  First pass duration (FPD)} \label{fig:timing2}
    \end{subfigure}
    \caption{Pearson correlations between model performance (Mean Absolute Error), and the deciles of the respective text characteristics. For measures with negative correlation coefficients, model performance increased with higher values of the text characteristics.}
    \label{fig:moreImp}
\end{figure*}

\end{document}